\documentclass[10pt,letterpaper,twocolumn]{article}
\usepackage[T1]{fontenc}
\usepackage{newtxtext}
\usepackage{helvet}
\usepackage{courier}
\usepackage[hmargin=0.75in,top=0.75in,bottom=1.25in]{geometry}
\usepackage[hyphens]{url}
\usepackage{graphicx}
\usepackage[numbers,sort&compress]{natbib}
\usepackage{caption}
\usepackage{placeins}
\usepackage{booktabs}
\usepackage{pifont}
\usepackage{xcolor}
\usepackage{tikz}
\usepackage[hidelinks]{hyperref}
\usetikzlibrary{positioning, shapes.geometric, arrows.meta}
\definecolor{memfill}{RGB}{226,239,249}
\definecolor{memline}{RGB}{45,114,178}
\definecolor{appfill}{RGB}{252,236,217}
\definecolor{appline}{RGB}{198,118,38}
\definecolor{neutfill}{RGB}{237,237,240}
\definecolor{neutline}{RGB}{88,88,94}
\newcommand{\cmark}{\textcolor{green!55!black}{\ding{51}}}
\newcommand{\xmark}{\textcolor{red!70!black}{\ding{55}}}

\newcommand{\pmarkn}[1]{\textcolor{orange!85!black}{$\sim^{#1}$}}

\title{\textbf{DolphinBench: Mapping the Pareto Frontier of Agent Memory}}
\author{Soumil Rathi, Deshraj Yadav, Taranjeet Singh\\Mem0\thanks{Correspondence: \texttt{dolphinbench@mem0.ai}}}
\date{}

\begin{document}

\maketitle

\begin{abstract}
\small
Agents today often take real world actions which depend on long term memory and context recall over time. However, most current memory benchmarks are built for a conversational question-answer format, where the question itself signals that some fact must be retrieved, and often which one. Moreover, benchmarks rarely require anything beyond accuracy from submissions, allowing memory systems to make unreasonable cost/time tradeoffs to achieve higher scores. We present DolphinBench, a benchmark that evaluates memory directly through an agent's task completion. DolphinBench includes three knowledge-work personas with roughly 500k tokens of user messages per persona and evaluates agents on tasks which depend on information from that history. We verify all 200 tasks per persona by running an agent with and without the relevant history, requiring success with it and failure without it. Finally, we require all evaluations to report total cost and latency alongside accuracy, which enables us to evaluate agent memory systems holistically. No existing memory benchmark combines all three. The dataset and evaluation code are available at \url{https://dolphinbench.ai}.
\end{abstract}

\section{Introduction}

The world is rapidly adopting agents in real world tasks, from software engineering \citep{swebench} and web tasks \citep{webarena} to customer service \citep{taubench} and long running projects \citep{wang2024agents}. Such agents are increasingly required to keep track of information across project histories and past sessions that outgrow their context windows, which memory systems \citep{memgpt,mem0,zep} are built to manage. It is important to evaluate this capability with memory benchmarks.

Memory benchmarks have largely not adapted to this setting. The dominant evaluation format is question answering \citep{memorybank,locomo,longmemeval}, where a system ingests a long history and answers questions about it. Such evaluations are inherently biased though, as a direct question acts as a signal to the system that some fact must be retrieved, and frequently mentions which fact. A QA benchmark asks ``What messaging platform does the team use?'', which already announces that a platform fact exists and is wanted; an action-based test of the same fact instead instructs the agent to post an update in the team's channel, and never mentions platforms at all. So, the evaluation effectively begins only after the most difficult step, recognizing that retrieval is required at all, has been completed on the system's behalf. An agent acting under an instruction may get a weaker hint, but nothing in the format guarantees one. Evaluating memory for agents requires removing the question and using memory-dependent tasks as the accuracy signal.

Memory benchmarks commonly run into two further limitations, the first of which is in the range of metrics demanded from a valid solution. Since memory and information retrieval is an optimization problem where tradeoffs can be chosen, a full evaluation requires comparing accuracy, cost, and latency. A system can achieve high accuracy, for instance, by re-reading the entire history at each query with frontier models, incurring significant latency and compute costs that a production deployment could not sustain. We therefore require every result to report total cost and median task latency alongside accuracy.

Second, because most benchmarks are generated synthetically and at scale, they are prone to defects: incorrect answer keys, mislabeled supporting evidence, and questions that are unanswerable even given the complete history \citep{locomoaudit,longmemevalissues}. Most benchmarks do not establish that each of their tests is actually solvable by a capable agent with perfect memory, which imposes an unknown ceiling on every reported score in case of such defects. We therefore verify every DolphinBench test using an agent with oracle memory: we provide the relevant history and require the agent to complete the task, then withhold that history and require it to fail. This checks that the task can be completed and that withholding the history affects success.

We introduce DolphinBench, a benchmark that evaluates agent memory through real world simulated actions. DolphinBench defines three knowledge-work personas: a startup CEO, an infrastructure engineer, and a product manager. We then simulate years of conversation (user input and tasks) for each persona with their agent. An agent processes this history and is then issued tasks whose correct actions depend on specific pieces of information within the session history. 

Together, these choices give DolphinBench three properties that existing memory benchmarks do not combine: tests that evaluate memory through action rather than recall, results that report cost and latency alongside accuracy, and verification of every test through agent runs with and without the relevant history. Across the evaluated models and harnesses, the highest-scoring agent completes 70.67\% of tasks. The memory systems rank differently across agents, and the most expensive agent is neither the most accurate nor the fastest.

\section{Related Work}

We identify three properties that are important for a memory benchmark for agents which mitigate the key issues faced by such benchmarks in practice. 
\begin{itemize}
    \item The test format must require action rather than pure text responses in a question-answer format.
    \item The final evaluation results must include cost and latency alongside accuracy.
    \item Each test should be solvable by the agent once it has perfect memory, and unsolvable without it (i.e. no ground truth errors, no hidden assumptions, and no tests that pass regardless of memory).
\end{itemize}

Table~\ref{tab:related_work} shows how prior memory benchmarks compare to DolphinBench on each property. The following subsections develop on the justification and existing work for each principle in turn.

\begin{table*}[t]
\centering\small
\begin{tabular}{lccc}
\toprule
Benchmark & Action-based & Cost+latency & Solvability-certified \\
\midrule
\multicolumn{4}{l}{\emph{Non-action paradigms}} \\
\quad MemoryBank \citep{memorybank} & \xmark & \xmark & \xmark \\
\quad LoCoMo \citep{locomo} & \xmark & \xmark & \xmark \\
\quad LongMemEval \citep{longmemeval} & \xmark & \xmark & \xmark \\
\quad MemoryAgentBench \citep{memoryagentbench} & \xmark & \pmarkn{a} & \xmark \\
\quad MemoryBench \citep{memorybench} & \xmark & \pmarkn{b} & \xmark \\
\quad MemBench \citep{membench} & \xmark & \pmarkn{a} & \xmark \\
\quad BEAM \citep{beam} & \xmark & \xmark & \xmark \\
\quad RealMem \citep{realmem} & \xmark & \pmarkn{f} & \xmark \\
\quad RealTalk \citep{realtalk} & \xmark & \xmark & \xmark \\
\quad MemProbe \citep{memprobe} & \xmark & \pmarkn{c} & \pmarkn{e} \\
\midrule
\multicolumn{4}{l}{\emph{Action-based paradigm}} \\
\quad MemoryCode \citep{memorycode} & \cmark & \xmark & \xmark \\
\quad StoryBench \citep{storybench} & \pmarkn{d} & \pmarkn{f} & \xmark \\
\quad Mem2ActBench \citep{mem2actbench} & \cmark & \xmark & \xmark \\
\quad STATE-Bench \citep{statebench} & \cmark & \pmarkn{g} & \xmark \\
\quad MemoryArena \citep{memoryarena} & \cmark & \pmarkn{a} & \xmark \\
\quad MEMTRACK \citep{memtrack} & \cmark & \xmark & \xmark \\
\midrule
\textbf{DolphinBench (ours)} & \cmark & \cmark & \cmark \\
\bottomrule
\end{tabular}
\caption{Memory benchmarks compared on the three properties an action-based memory evaluation requires. \cmark: yes, \xmark: no, $\sim$: partial: $^{a}$latency but no cost; $^{b}$latency only in an appendix; $^{c}$token counts as the only cost signal, no latency; $^{d}$actions in a story/text-game setting rather than app state; $^{e}$checks task design and information disclosure after evaluation; $^{f}$timing and token counts, but no monetary cost; $^{g}$monetary cost but no latency.}
\label{tab:related_work}
\end{table*}

\subsection{Task Format}

Memory evaluation began as conversational QA \citep{memorybank,locomo}, later expanding into categories like temporal reasoning, knowledge updates, and abstention \citep{longmemeval,memoryagentbench} while keeping the QA format. The format carries the retrieval signal discussed in the introduction, and an action-style task removes it: the agent has to ``know to remember'', which makes such tasks both harder and more representative of real world usage.

Some recent benchmarks have moved toward action-based testing. Mem2ActBench \citep{mem2actbench} scores agents on task completion in simulated environments; MemoryArena \citep{memoryarena} evaluates interdependent agentic subtasks; STATE-Bench \citep{statebench} tests stateful agent actions; MEMTRACK \citep{memtrack} tracks state across interleaved Slack, Linear, and Git timelines.

\subsection{Metrics}

Most benchmarks do not require cost and latency, making the partial reporting on these benchmarks inconsistent. For example, STATE-Bench reports dollar cost per task but not latency; MemoryArena reports latency but no aggregate cost; MemoryAgentBench reports memory construction and query execution times but not monetary cost. However, neither cost nor latency alone is enough, since inference choices like batch size and hardware trade them off. Thus, both these metrics should appear in any valid submission.

Accuracy alone is also easy to skew: the same benchmark can produce very different numbers under a modified judge prompt or judge model, with \citet{locomoaudit} showing that one evaluated judge configuration accepted approximately 63\% of deliberately incorrect, topically related answers. Cost and latency are harder to quietly improve, which makes them an anchor for comparisons across papers.

\subsection{Validity}
\label{sec:validity} 
An independent audit of LoCoMo \citep{locomoaudit} reported incorrect answers for 99 of 1,540 non-adversarial questions (6.4\%). Other popular benchmarks contain similar issues. LongMemEval contains ground-truth arithmetic errors and mislabeled evidence, surfaced as GitHub issues by the community \citep{longmemevalissues}.

Careful pipeline construction reduces error rates but does not eliminate them, so BEAM \citep{beam} adds human review, checking that every question is answerable. However, a test can pass that review and still be solvable without any memory, and then every system gets it right whether its memory works or not. Every DolphinBench test therefore has to fail without the relevant sessions as well as pass with them. We verify both automatically before a test ships, as described in the Test Construction section.

\section{DolphinBench}

DolphinBench is a benchmark built to evaluate agents on their memory which directly tests task success. The benchmark covers three personas: Morgan, a startup CEO; Alex, an infrastructure engineer; and Riley, a product manager. These were chosen to have a wide coverage of different tools/apps the agent would use, queries across different task types, and thus different memory access patterns as well. 

Each persona has a simulated history spanning several years, with 3,400--5,128 user messages totalling approximately 500k tokens\footnote{Measured with tiktoken's \texttt{o200k\_base}, over user-message content only.}. The agents go through this simulated history, after which they are tested on tasks that require information from within those simulated sessions. The benchmark contains 600 tests, 200 per persona. Tool calls and apps (Notion, Gmail, GitHub, etc.) are both simulated to reproducibly test real world actions \citep{taubench,webarena}.

\paragraph{Worked example.} Morgan test 001:\\[2pt]
\textbf{History.} In a January session, Morgan mentions a coffee preference: \textit{``still doing blue bottle if i'm out early enough.''} \\
\textbf{Instruction.} \textit{``I'm out early. Please order one small latte for pickup.''}\\
\textbf{Answer \& grading.} The agent must use \texttt{place\_order} to order from Blue Bottle. A deterministic check requires the order to succeed, and an LLM judge checks that the restaurant is Blue Bottle.

\noindent The request specifies the drink, size, and pickup, but not the cafe. The agent must use Morgan's earlier preference to choose where to place the order.

\subsection{Construction}

\begin{figure*}[t]
\centering
\resizebox{\textwidth}{!}{%
\begin{tikzpicture}[
  >={Stealth[round]},
  font=\small,
  box/.style={rectangle, rounded corners=4pt, draw, line width=0.8pt, align=center,
              inner xsep=7pt, inner ysep=6pt, minimum height=1.15cm, minimum width=2.9cm},
  neut/.style={box, draw=neutline, fill=neutfill},
  mem/.style={box, draw=memline, fill=memfill},
  app/.style={box, draw=appline, fill=appfill},
  ship/.style={box, draw=green!45!black, fill=green!14, minimum width=1.7cm},
  flow/.style={->, line width=0.9pt, draw=black!60, shorten >=2pt, shorten <=2pt},
  redo/.style={->, line width=0.8pt, draw=black!45, dashed, shorten >=2pt, shorten <=2pt},
]
\node[neut] (spec) at (0,0)       {\textbf{Personas}\\[1pt]{\scriptsize\itshape\color{black!60} people, apps, voice}};
\node[neut] (plan) at (4.6,0)     {\textbf{History planning}\\[1pt]{\scriptsize\itshape\color{black!60} overall narrative,}\\{\scriptsize\itshape\color{black!60} quarters, weeks}};
\node[app]  (env)  at (9.2,0)     {\textbf{App records}\\[1pt]{\scriptsize\itshape\color{black!60} what the tools serve}};
\node[mem]  (conv) at (13.8,0)    {\textbf{User messages}\\[1pt]{\scriptsize\itshape\color{black!60} what the agent ingests}};
\node[neut] (test) at (9.2,-3.4)  {\textbf{Test construction}\\[1pt]{\scriptsize\itshape\color{black!60} task + grading checks}};
\node[neut] (gate) at (4.6,-3.4)  {\textbf{Test verification}\\[1pt]{\scriptsize\itshape\color{black!60} 2$\times$ pass with oracle messages}\\{\scriptsize\itshape\color{black!60} 2$\times$ fail without}};
\node[ship] (ship) at (0,-3.4)    {\textbf{Accepted tests}};
\draw[flow] (spec) -- (plan);
\draw[flow] (plan) -- (env);
\draw[flow] (env) -- (conv);
\draw[flow] (plan.north) -- ++(0,0.9) -| (conv.north);
\draw[flow] (conv.south) |- node[above, align=center, font=\scriptsize, pos=0.75]{facts and\\source messages} (test.east);
\draw[flow] (env.south) -- (test.north);
\draw[flow] (test.west) -- (gate.east);
\draw[flow] (gate.west) -- (ship.east);
\draw[redo] (gate.south) to[out=-90,in=-90, looseness=0.55]
  node[below, font=\scriptsize, text=black!60, pos=0.5]{revise and recheck}
  (test.south);
\end{tikzpicture}}
\caption{The DolphinBench construction pipeline.}
\label{fig:construction}
\end{figure*}
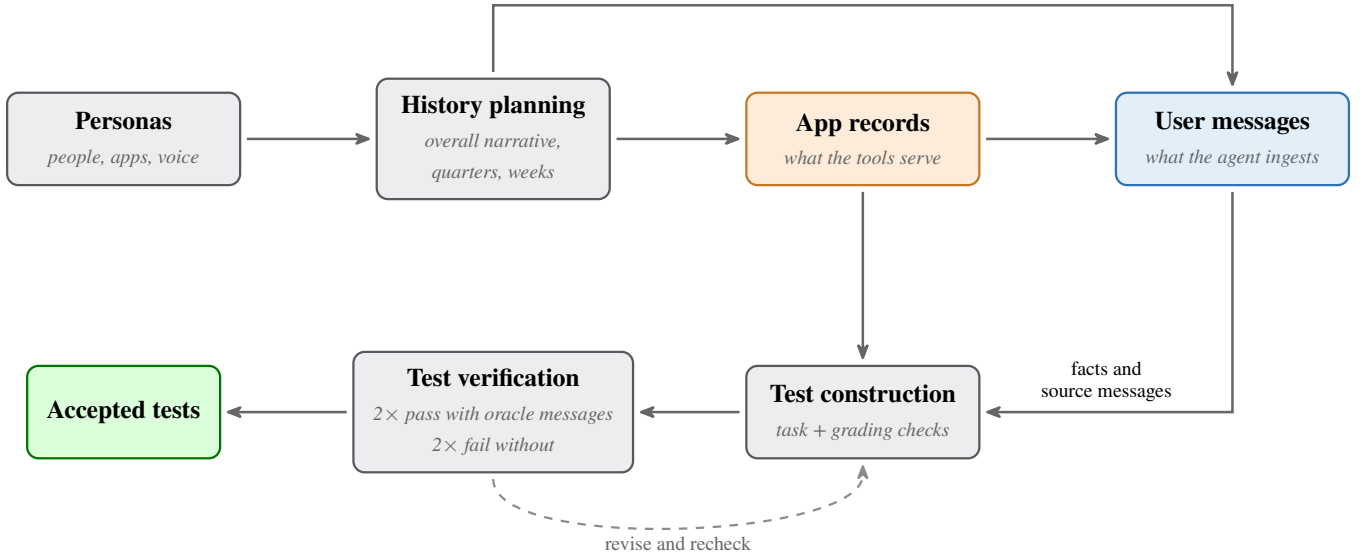

DolphinBench is built in hierarchical stages (Figure~\ref{fig:construction}), an approach also used by BEAM \citep{beam}. We first outline each persona's history over several years, then develop it into quarterly plans, weekly events, and conversations. The resulting messages and app records provide the basis for test construction.

\subsubsection{Personas}

We define each persona's name, role, relationships, and tools, including calendar, email, CRM, and deployment tools. An LLM drafts these descriptions, which later stages use when planning events and writing messages. We update the definitions as people or tools change throughout the history. DolphinBench's three personas are Morgan, a startup CEO who manages investors, hiring, board communication, and strategy; Alex, an infrastructure engineer who handles deploys, incidents, on-call rotations, and runbooks; and Riley, a product manager who runs experiments, funnels, and retention work.

\subsubsection{History Planning}

We outline the persona's work and personal life over several years. An LLM develops this narrative into quarterly plans, specifying dated events and the facts they are intended to establish or change. These include decisions, preferences, and developments in ongoing projects and relationships. We review the plans before generating the corresponding history.

An LLM then develops each quarter into weekly events and conversations. For each planned conversation, it specifies a date, why the persona contacts the agent, and what they ask for or share. The plan includes everyday exchanges alongside the quarter's main events.

\subsubsection{History Generation}

Before writing these conversations, we update the simulated apps to reflect the planned events. An LLM specifies the required tool calls and arguments, such as updating an event in Google Calendar. We execute these calls in chronological order so that the messages can refer to the resulting app records.

The model writing the messages receives these records and tool outputs, together with the weekly plan, relevant facts, and earlier messages. It writes the user messages in the persona's voice, using this context to keep them consistent with the apps and preceding conversations. Each released history ends at the first complete session that brings it to at least 500k user-message tokens.

We author only the user side of each conversation, because each agent would reply differently. The agent produces its own replies during ingestion. QA benchmarks instead provide a fixed transcript of both sides \citep{locomo,longmemeval}, which fits a system that reads a conversation but not one that participates in it.

We record which facts each message mentions and which it introduces or changes. Before using a fact to construct a test, we check its description against the source messages and correct any claims they do not support.

\subsubsection{Test Construction}

An LLM proposes tasks that require the agent to use earlier information to complete new work. To design these tasks, it receives facts from the history, descriptions of the available tools, the evaluation date, and summaries of previously proposed tasks. For each task, it specifies the required facts, the intended actions, and what would be missing or wrong without the history. Each selected fact must affect a necessary part of the completed work.

An LLM then receives the proposed task, relevant source messages and later updates, available tools, and app records. It writes the user request and grading checks. The request supplies the information needed to act while leaving the remembered details for the agent to determine. We avoid cues that announce a stored preference, such as ``the way I usually do it''. Checks on remembered information cite the supporting source messages and require only the details needed for the task.

We select the app records needed to carry out the task and add present-day records where the proposed situation requires them. These records supply ordinary task inputs without revealing the remembered answer. We review the request, app records, and grading checks against the source messages before verification.

\subsubsection{Test Verification}
\label{oracle}
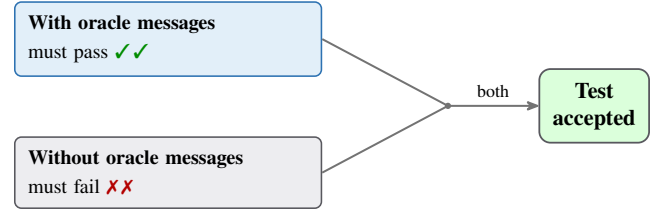
\begin{figure}[t]
\centering
\resizebox{\columnwidth}{!}{%
\begin{tikzpicture}[
  >={Stealth[round]},
  font=\small,
  cond/.style={rectangle, rounded corners=3pt, draw, line width=0.7pt, align=left, inner sep=6pt, text width=4.4cm},
  withf/.style={cond, draw=memline, fill=memfill},
  nomem/.style={cond, draw=neutline, fill=neutfill},
  ship/.style={rectangle, rounded corners=5pt, draw=neutline, line width=0.8pt, fill=green!14, align=center, inner sep=6pt, font=\bfseries},
  link/.style={line width=0.9pt, draw=black!55},
]
\node[withf] (w) at (0,1.05) {\textbf{With oracle messages}\\[2pt] must pass \cmark\,\cmark};
\node[nomem] (n) at (0,-1.05) {\textbf{Without oracle messages}\\[2pt] must fail \xmark\,\xmark};
\coordinate (J) at (4.4,0);
\node[ship] (s) at (6.7,0) {Test\\accepted};
\draw[link] (w.east) -- (J);
\draw[link] (n.east) -- (J);
\fill[black!55] (J) circle (1.4pt);
\draw[link,->] (J) -- node[above, font=\footnotesize]{both} (s.west);
\end{tikzpicture}}
\caption{Test verification. The agent must pass both runs with oracle messages and fail both runs without them.}
\label{fig:oracle}
\end{figure}

For each test, we provide the original history messages needed to complete it as oracle messages. We run the test twice with GPT-5.6-Luna given these oracle messages, and twice without them (Figure~\ref{fig:oracle}). The task, tools, and starting app records remain the same. Both runs with history must pass every grading check, and both runs without history must fail.

We inspect the runs to check that failure without history reflects missing information rather than tool errors or defects in the test. We revise candidates that fail these checks and exclude those that do not meet the acceptance conditions.

\begin{table*}[t]
\centering\small
\setlength{\tabcolsep}{5pt}
\begin{tabular}{lllrrrrr}
\toprule
& & & & \multicolumn{3}{c}{Cost (\$)} & \\
\cmidrule(lr){5-7}
Harness & Agent model & Memory system & Accuracy (\%)
& Agent & Memory system & Total & Median latency (s) \\
\midrule
Hermes & GPT-5.6-Luna & Built-in memory
& 65.67 & 61.48 & 0.00 & \textbf{61.48} & 44.35 \\
 & & Mem0
& \textbf{70.67} & 61.54 & 34.68 & 96.21 & \textbf{37.69} \\
 & & Honcho
& 68.50 & 96.56 & 46.30 & 142.85 & 44.66 \\
& & Hindsight
& 69.50 & 57.99 & 26.66 & 84.65 & 55.31 \\
 & & Supermemory
& 59.17 & 63.86 & 281.79 & 345.64 & 52.68 \\
\midrule
Hermes & MiniMax M3 & Built-in memory
& 26.50 & 107.03 & 0.00 & \textbf{107.03} & 51.40 \\
 & & Mem0
& \textbf{47.83} & 111.71 & 36.31 & 148.02 & \textbf{46.95} \\
 & & Honcho
& 47.33 & 168.39 & 43.46 & 211.85 & 80.33 \\
& & Hindsight
& 41.00 & 119.51 & 38.82 & 158.33 & 144.28 \\
 & & Supermemory
& 20.00 & 111.68 & 386.02 & 497.70 & 137.52 \\
\midrule
Claude Code & Claude Sonnet 5 & Built-in memory
& 26.33 & 1,132.75 & 0.00 & \textbf{1,132.75} & \textbf{32.19} \\
 & & Mem0
& 32.33 & 1,803.63 & 26.94 & 1,830.57 & 42.30 \\
 & & Honcho
& \textbf{35.83} & 1,556.02 & 9.79 & 1,565.82 & 37.92 \\
\bottomrule
\end{tabular}
\caption{Results on DolphinBench: comparing the accuracy, cost, and latency of agents using different memory systems across 600 tasks spanning three personas. Bold marks the highest accuracy, lowest total cost, and lowest median latency within each harness--model group.}
\label{tab:results}
\end{table*}

\subsection{Grading \& Metrics}
\label{sec:metrics} 
We grade task completion by checking the agent's tool calls against the actions and results required by the task. During test construction, we choose how to check each requirement:
\begin{itemize}
    \item We use an LLM judge \citep{llmjudge} (GPT-5.6-Sol) when checking a requirement requires interpreting meaning, such as whether an email conveys a particular decision. The judge receives the user's request and the arguments of the tool call being checked, and is instructed to accept equivalent wording.
    \item We use deterministic checks when a requirement can be tested directly, such as whether a required tool was called or whether an identifier, number, date, or list matches the required value.
\end{itemize}

The checks must establish that each required action was completed correctly. For example, an email must have both the correct recipient and the required content in the same send call; satisfying those checks across different emails does not count. If a task requires two separate emails, one send call cannot count as both. A task passes only when every check passes.

Every submission must report accuracy, total cost across ingestion and testing, and median task latency. We calculate accuracy as the fraction of tasks passed and take the median of the individual test latencies across all personas.

\section{Results}

We evaluate agents with different combinations of harnesses, language models, and memory systems. Table~\ref{tab:results} reports the accuracy, cost, and latency of each combination.

Accuracy is the percentage of tasks passed. Total cost covers ingestion and testing, including memory processing. Latency is the median time per test, including tool use.

\nocite{supermemory}

Memory performance reflects the capabilities of the model, the harness, and the memory system working together. With Hermes \citep{hermes} and Mem0 \citep{mem0}, accuracy reaches 70.67\% with GPT-5.6-Luna, compared with 47.83\% with MiniMax M3. The memory systems also rank differently across configurations: Mem0 scores highest with both Hermes models, while Honcho \citep{honcho} scores highest with Claude Code \citep{claudecode} and Claude Sonnet 5.

Higher accuracy need not require longer response times. With Hermes and GPT-5.6-Luna, Mem0 improves accuracy over built-in memory by five percentage points and reduces median task latency from 44.35 to 37.69 seconds, while total cost rises from \$61.48 to \$96.21.

Lower cost and higher accuracy can also come with longer response times. With the same harness and model, Hindsight \citep{hindsight} costs \$84.65 compared with Honcho's \$142.85 and scores one percentage point higher, but its median task latency is 55.31 seconds compared with Honcho's 44.66 seconds.

Evaluating each configuration on all three metrics lets DolphinBench map the Pareto frontier and helps users choose the configuration that best matches their accuracy, cost, and latency requirements.

We release the configuration, cost calculations, and full trace of every run in Table~\ref{tab:results} (tool calls, arguments, and per-check verdicts) alongside the dataset.

\section{Limitations}
\label{sec:limitations}

\paragraph{Synthetic data.} We generate conversation histories rather than collect them from real users. These histories may not capture the variety of writing styles, languages, and interaction patterns found in real conversations. Checking facts against source messages and verifying the tests helps catch errors, but inconsistencies may remain.

\paragraph{Domain coverage.} Three personas cover only a small part of knowledge work. They exercise different tools and memory requirements, but do not represent domains such as law, medicine, or long-term customer support. The construction pipeline supports adding personas to broaden this coverage.

\paragraph{Simulated environment.} Agents interact with simulated apps rather than live services. The apps support selected operations and errors, but do not reproduce the full behavior of production systems, including service outages and changes made by other users during a task. This makes the tasks reproducible while leaving some difficulties of real deployments untested.

\paragraph{Task length.} Each task requires one to four separately graded actions, though agents may make additional calls to find information or correct mistakes. These tasks test whether agents use remembered information correctly without requiring long workflows. Longer tasks would extend the evaluation to using memory across many dependent steps.

\paragraph{Separate ingestion and testing.} Agents process the full history before taking any tests. This lets us compare agents against the same completed history, but does not test how they handle new conversations and tasks arriving throughout an ongoing evaluation. Interleaving ingestion and testing would better represent that continuous use.

\section{Conclusion}

DolphinBench evaluates agent memory through the actions agents take. We grade task completion, verify every test through agent runs with and without the relevant history, and require every result to report cost and latency alongside accuracy.

Our results show that a more accurate agent need not be slower, and a more expensive agent need not be more accurate. We therefore compare accuracy, cost, and latency together to assess each agent configuration as a whole.

We intend to keep the benchmark growing: histories that exceed model context windows, more personas, longer tasks, and interleaved ingestion and testing that better matches real deployments. As agents improve, the benchmarks should become harder with them.

\FloatBarrier
\begingroup
\small
\raggedright
\bibliographystyle{plainnat}
\bibliography{dolphinbench}
\endgroup

\end{document}